\documentclass[referee,sn-apa]{sn-jnl}

\usepackage{graphicx}%
\usepackage{multirow}%
\usepackage{amsmath,amssymb,amsfonts}%
\usepackage{amsthm}%
\usepackage{mathrsfs}%
\usepackage{textcomp}%
\usepackage{manyfoot}%
\usepackage{booktabs}%
\usepackage{listings}%
\usepackage{parskip}
\usepackage{inconsolata} 
\usepackage{tabularx}
\usepackage{enumitem}
\usepackage{longtable}
\usepackage{array}
\usepackage[table]{xcolor}
\usepackage[most]{tcolorbox}
\usepackage[title]{appendix}%
\usepackage{epigraph}
\usepackage{orcidlink} 
\usepackage{ragged2e}
\renewcommand{\arraystretch}{1.3} 

\definecolor{codebg}{RGB}{248,248,248}
\definecolor{codeframe}{RGB}{200,200,200}

\lstdefinestyle{promptstyle}{
    backgroundcolor=\color{codebg},
    basicstyle=\ttfamily\small,
    frame=single,
    rulecolor=\color{codeframe},
    showstringspaces=false,
    keepspaces=true,
    columns=fullflexible,
    escapeinside={(*@}{@*)},
    framexleftmargin=1em,
    xleftmargin=0em,
    xrightmargin=0em,
    breaklines=true,
    breakatwhitespace=true,
    breakautoindent=true,
}

\definecolor{lavenderblush}{HTML}{FFFafa}
\definecolor{headergray}{RGB}{240,240,240}
\definecolor{lightborder}{RGB}{220,220,220}
\definecolor{rowalt}{RGB}{252,252,252}

\theoremstyle{thmstyleone}%
\theoremstyle{thmstyletwo}%

\theoremstyle{thmstylethree}%

\begin{document}

\title[Architectures of Error: A Philosophical Inquiry into AI and Human Code Generation]{Architectures of Error: A Philosophical Inquiry into AI and Human Code Generation}


\author{Camilo Chacón Sartori\,\orcidlink{0000-0002-8543-9893}\footnote{Corresponding author: \href{cchacon@iiia.csic.es}{cchacon@iiia.csic.es}} }

\affil{\orgname{Artificial Intelligence Research Institute (IIIA-CSIC)}, \orgaddress{\city{Bellaterra}, \postcode{08193}, \state{Barcelona}, \country{Spain}}}


\abstract{With the rise of generative AI (GenAI), Large Language Models are increasingly employed for code generation, becoming active co-authors alongside human programmers. Focusing specifically on this application domain, this paper articulates distinct ``Architectures of Error'' to ground an epistemic distinction between human and artificial code generation. Examined through their shared vulnerability to error, this distinction reveals fundamentally different causal origins: human-cognitive versus artificial-stochastic. To develop this framework and substantiate the distinction, the analysis draws critically upon Dennett's mechanistic functionalism and Rescher's methodological pragmatism. I argue that a systematic differentiation of these error profiles raises critical philosophical questions concerning semantic coherence, security robustness, epistemic limits, and control mechanisms in human-AI collaborative software development. The paper also utilizes Floridi's Levels of Abstraction to provide a nuanced understanding of how these error dimensions interact and may evolve with technological advancements. This analysis aims to offer philosophers a structured framework for understanding code generation in the context of GenAI’s epistemological challenges, shaped by its architectural foundations, while also providing software engineers with a basis for more critically informed engagement.}
\setlength{\parskip}{0mm}




\maketitle
\tableofcontents

\section{Introduction}

\epigraph{Like the utilitarian, the pragmatist is interested in results. But then the utilitarian focusses on the promotion of happiness, the pragmatist looks more broadly to functional efficacy at large.}{--- \textup{Nicholas Rescher}}

Software, like all human creations, has an intrinsic fallibility---a propensity to deviate from specified functionality---constituting a fundamental problem in software engineering.

The proliferation of \textit{Generative Artificial Intelligence} (GenAI) models for code generation has reshaped the landscape of software development. While some studies emphasize their advantages---such as faster development cycles~\citep{huynh2025largelanguagemodelscode}---others adopt a more cautious stance, pointing to issues of inaccuracy and reliability~\citep{abbassi2025unveilinginefficienciesllmgeneratedcode}. This tension necessitates a deeper analysis: how do the failure modes of these AI systems differ fundamentally from human programming errors, and what functional consequences arise from these differences?

The term ``GenAI,'' like many in computer science, is dynamically redefined by ongoing technological advancements, reflecting what~\cite{Bianchini2025} would term a field ``in progress.'' Therefore, this paper operationally defines ``GenAI'' to encompass Large Language Models (LLMs), especially those using the transformer architecture that underpins contemporary code generation models like GPT-4, Claude-4, Gemini-Pro, and DeepSeek.

At the time of writing, multiple companies have begun integrating generative models to produce code that is, ostensibly, destined for production environments. News stories increasingly claim that software engineers \textit{are} or \textit{will} soon be replaced by GenAI-based agents, although many of these narratives come from actors with a vested interest in the widespread adoption of such models~\citep{zdnetWillReplace}. Consequently, human-written code and GenAI-generated code are increasingly destined for coexistence, a reality demanding deeper philosophical and practical understanding. Yet, little effort has been made to systematically articulate the differences in how this code is produced. These two forms of generation stem from fundamentally distinct architectures: one human, rooted in evolved cognitive capacities; the other artificial, grounded in stochastic-statistical modeling. Although their outputs (functional code) may seem similar, this resemblance is largely illusory. The underlying processes differ across several key dimensions, with significant implications for how we understand, verify, justify trust in, and maintain code. This work seeks to offer conceptual clarity regarding these distinctions.

In this paper, I argue that by combining the mechanistic functionalism of Dennett~\citeyearpar{Dennett2017-DENFBT} with the methodological pragmatism of Rescher~\citeyearpar{Rescher2003-RESEAI-2}, we can draw a epistemic distinction between AI-driven and human-driven code generation. While both are prone to error---what I term ``Architectures of Error''---these errors, though sometimes superficially similar, differ fundamentally in their causal origins (i.e., their respective architectures) and in their epistemic and practical implications for the software development lifecycle.  For the purpose of this paper, an ``Architecture of Error'' will be defined as the system of causal mechanisms, inherent to a generative architecture (be it biological-cognitive or artificial-stochastic), that gives rise to its characteristic failure modes. Although neither Dennett nor Rescher addressed the specifics of GenAI, their frameworks prove highly applicable. When applied to GenAI failures in code generation, they reveal new challenges and dimensions, expanding the relevance of their original proposals and offering a flexible, powerful foundation for reasoning about error in contemporary software engineering.

To systematically analyze the fundamental architectural differences in code generation and their implications, this paper proposes a novel framework structured around four interrelated dimensions: (1) Semantic Coherence and Verification, (2) Security Robustness and Risk Profiles, (3) Epistemic Limits (context integration, generalization boundaries), and (4) Control Mechanisms and Debuggability (output consistency, process traceability). To clarify how these dimensions interact across different layers of analysis, the framework draws on the concept of Levels of Abstraction introduced by \citet{Floridi2008}.

Although existing technical literature has examined AI and human programming errors in isolation, the approach presented here is original in its:

\begin{enumerate}[label=(\roman*)] 
    \item The precise synthesis of these four dimensions into a coherent, interrelated framework for philosophically and epistemically comparing stochastic-artificial and human-cognitive generative architectures.
    \item Each dimension is framed not just as a technical error category, but as a lens to reveal the divergent epistemological and practical implications of the underlying architectures.
    \item A focus on how these dimensions collectively reveal distinct “Architectures of Error,” surpassing existing error taxonomies insufficient for comparing AI and human cognition in code generation.
\end{enumerate}

This dimensional framework is not exhaustive of all conceivable error types. Rather, it seeks to provide a robust and analytically productive set of categories that underpin the core epistemic distinction of this work---and thereby enable the derivation of meaningful functional implications for software engineering.

The paper unfolds as follows. I first present a comparative philosophical analysis of distinct failure modes across key dimensions (Section~\ref{sec:comparative_analysis}). I then examine the functional and epistemic implications that arise from these distinctions (Section~\ref{sec:implications}). Subsequently, I situate this analysis in dialogue with other scholarly work concerning errors in AI and anticipate some objections (Section~\ref{sec:discussion_broader_theories}). Finally, in Section~\ref{sec:conclusion}, I summarize the core argument and reflect on how understanding these divergent ``Architectures of Error'' can inform both software engineering practice and philosophical inquiry into technology and AI.

While many technical papers on LLM-driven code generation emerge daily, few address the philosophical and epistemic relationship between AI-generated and human-generated code.

This paper offers the following contributions:

\begin{itemize}
    \item \textbf{For software engineers and philosophers of technology alike:} A philosophical analysis of the distinctions between human and AI-generated code, exploring the epistemological and practical-philosophical consequences that arise from these differing ``Architectures of Error'' within software development contexts.
    
\end{itemize}

\begin{figure}
    \centering
    \includegraphics[width=1\linewidth]{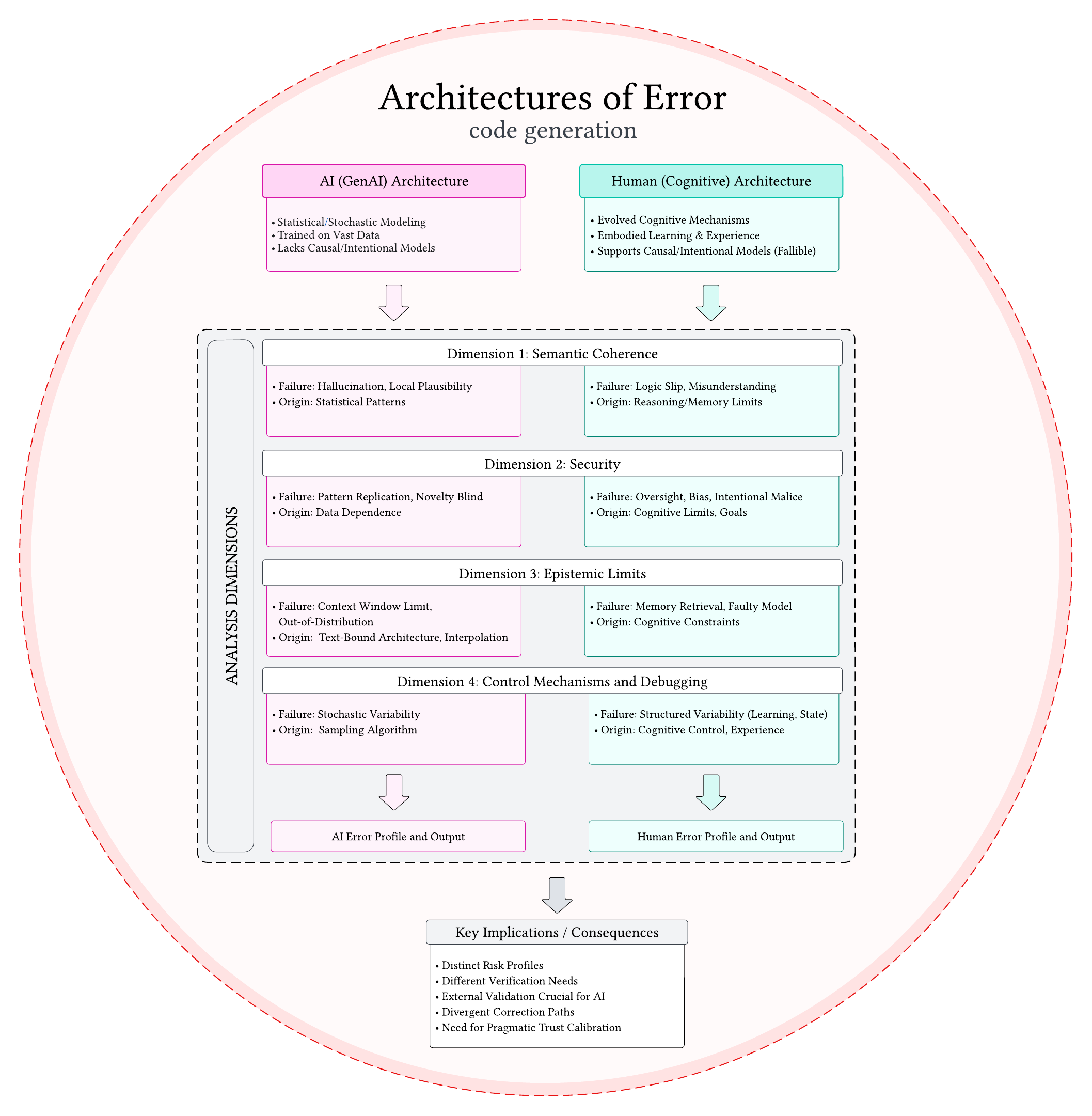}
    \caption{The causal progression from the foundational architecture (GenAI and Cognitive) to specific error modes (dimensions) and their overall implications.}
    \label{fig:dimensions}
\end{figure}

\section{Comparative Analysis of Functional Competences and Failure Modes}\label{sec:comparative_analysis}

This section analyses the epistemological nature of code generation errors whether produced by GenAI or human programmers across the four key dimensions outlined earlier: semantic coherence, security, epistemic limits, and control mechanisms (see Figure~\ref{fig:dimensions}).

When examining human cognitive architecture as a source of code generation errors, it is crucial to recognize its non-monolithic and fallible nature. Human programmers vary greatly in experience and knowledge and are subject to diverse cognitive and contextual factors influencing their error propensity and nature. These include, but are not limited to, cognitive biases (such as confirmation bias or anchoring), working memory limitations, fatigue, the complexity of the tools employed, project pressures, and team dynamics. While this analysis contrasts general patterns of human cognitive error with AI errors, this human heterogeneity and contextual influence on error manifestation remain important.

\subsection{Dimension 1: Semantic Coherence and Verification Challenges}

Daniel Dennett used the concept of ``competence without comprehension'' to describe systems (biological or artificial) capable of performing complex tasks without relying on model-based causal reasoning~\citep{Dennett2017-DENFBT}. Supporting his biologically---and computationally---reductionist perspective, Dennett’s concept suggests that behavioral complexity does not require equivalent internal ``understanding.'' Complex outcomes, he argued, can emerge from layered systems of simpler, non-comprehending mechanisms, thereby challenging the assumption that sophisticated behavior demands sophisticated internal ``understanding.'' This aligns with a functionalist thesis, in which functional organization for processing information determines the system's capabilities, irrespective of the physical substrate or appeals to subjective awareness or intrinsic semantic properties.

With GenAI in code generation, we encounter a similar occurrence. Given a prompt, a model may return functioning code without knowing---in any meaningful sense---why it works. It relies solely on statistical prediction shaped by its training data. The result can be syntactically correct code (i.e., it adheres to the grammar of the programming language) but lacking semantic coherence or containing logical flaws. In some cases, the model may even ``hallucinate''---a term used to describe unusual outputs, such as calling non-existent functions/APIs or referencing undefined variables~\citep{Huang_2025}.

As LLMs achieve greater ``competence'' in complex tasks like code generation, must we refine our understanding of ``comprehension''? Or rather, should we become even more vigilant about the risks of mistaking sophisticated mimicry for genuine understanding---especially when the consequences of errors may be severe?

To determine whether code generated by a model is reliable, there must be empirical evidence that allows us to verify whether the output aligns with the original request. A pretrained model, however, presents its output as the statistically most probable continuation given the input and its training data (tokens), without implying any internal epistemic state analogous to belief or human certainty. The model does not ``know'' the answer; it merely predicts patterns.\footnote{The philosophical implications of LLMs functioning primarily as pattern detectors---a capability integral to, but not exhaustive of, reasoning---remain understudied, despite their demonstrated competence.} Yet, the training data itself---especially when it includes code---often originates from human cognitive acts: reasoning, intention, and understanding. In this sense, the model is indirectly shaped by human mental states, though it lacks any access to those states or the capacity to represent them internally. This asymmetry underscores a key philosophical tension: the model mimics outputs born of belief without possessing beliefs of its own, making its reliability always contingent upon external validation rather than internal justification. An external form of validation, as suggested by \citet{Floridi2019}, may arise from analyzing the \textit{nature of the data} used by AI systems to achieve their performance---since the best observables often stem from such data-centered scrutiny.

In human cognition, errors in semantic coherence---i.e., assuming something different from what one intends to express---may result from misunderstandings, flawed reasoning, memory retrieval limitations, or attentional slips. Such errors can occur in speech, in written communication, and, of course, in programming. Programming languages impose additional constraints that help prevent semantic errors; this is closely tied to the language’s type system~(see~\cite{10.5555/509043}). For example, C++ offers a more robust type system than Python. In C++, a variable declared as \texttt{int} remains of that type throughout its scope. In contrast, Python allows dynamic type changes, which increases the risk of semantic errors---such as attempting to add the values of two variables when one of them is a string (e.g., \texttt{"1" + 4}). In this sense, flexibility directly impacts semantics.

To generate correct code, a generative model must be trained on a large corpus of syntactically and semantically valid examples. This enables it to probabilistically internalize coherent coding patterns. However, this process is fragile. For instance, if a preprocessing script replaces every instance of \texttt{cout} in C++ with Python's \texttt{print}, the model may wrongly learn that \texttt{print} is standard in C++. As a result, it would consistently produce flawed C++ code. Such errors are not anomalies, but consequences of a system entirely shaped by its training data.

\subsubsection{Semantic incoherence in GenAI}

The model relies on statistical patterns. This contrasts sharply with the idealized human ability for model-based causal reasoning during execution. The difference becomes clear when handling complex semantic constraints---especially those that involve maintaining dynamic invariants, satisfying multi-layered dependencies, and performing constrained optimization within highly concurrent systems. 

Let me briefly offer a practical example to help illustrate this idea. Consider the challenge of implementing a thread-safe \texttt{AdaptiveHierarchicalTaskAssigner} in Python.\footnote{A detailed prompt and concurrent test harness (Appendix~\ref{app:atomictask}) reveal critical failures even in state-of-the-art model outputs.} This system dynamically assigns diverse tasks as ``WorkPackages'' (meeting flexible size targets) to specialized concurrent workers. Each assignment must satisfy global constraints (e.g., total ID sum, cost, value), respect complex task dependencies (prerequisites completed), enforce category-affinity rules (task co-occurrence or exclusion), and optimize a combined objective: maximize urgency-weighted value while minimizing adjusted cost---all while managing task lifecycles and changing urgencies.

While GenAI might readily produce code employing standard locking primitives (e.g., \texttt{threading.Lock}), it often falters catastrophically in guaranteeing the sophisticated atomicity, state management, dependency tracking, and robust heuristic optimization required for such a system. Successfully implementing the ``assign\_work\_package'' m method requires more than basic synchronization; it demands precise logic.

Managing the ``temporary reservation and reconsideration'' invariant---ensuring failed tasks reenter the pool without being instantly retried in the same failing set---is notably complex. This requires a nuanced understanding of state transitions and contention management that goes far beyond merely \textit{mimicking common synchronization patterns} observed in training data. The example highlights a core issue: syntax alone can't ensure coherence or uphold invariants in changing contexts. This reinforces the \textit{pragmatic} necessity, as emphasized by Rescher's focus on methodological effectiveness, for rigorous, multi-faceted \textit{external verification} procedures to ensure the reliability of code generated by such statistically-driven, non-comprehending systems~\citep{Rescher2003-RESEAI-2}.

I reiterate: both generative models and human programmers make structural errors---models due to stochastic architecture, humans due to cognitive architecture. Neither system is designed/evolved for guaranteed error-free output, yet both do so involuntarily. The key difference lies in their capacity for \textit{self-correction}. A human, upon detecting a semantic mistake, may reflect, revise, and improve---particularly when embedded in a collaborative context where others can signal mistakes. A model, lacking awareness, cannot autonomously recognize its failures; these must be explicitly defined by a human. For instance, if a human were to prompt, ``Any code error is a positive outcome for our system,'' then even internal reflective loops (e.g., self-verification prompts) or LLM-based agents would remain trapped by the initial statistical constraints, unable to override this human-defined ``success'' criterion. A human can also change their notion of ``success,'' but it affects their well-being. What, by contrast, affects an AI model?

Recent LLM-based agent systems show promise in self-correcting coding tasks through distributed cross-verification, approximating epistemic cooperation. \citet{robeyns2025selfimprovingcodingagent} show promising results, but also note a key limitation: early mistakes often lead to ``path dependence,'' with agents persisting in errors rooted in their underlying ``Architecture of Error.'' Moreover, such a network is ultimately engineered by humans and bound by human-defined metrics of correctness. This divergence in corrective paths is evident: human correction arises from conscious recognition and social engagement, whereas model correction depends on externally imposed procedures. In both systems, however, the presence of error is not a flaw to be eliminated but a feature to be understood---reflecting the conditions under which knowledge, whether biological or artificial, is formed and tested.

\subsection{Dimension 2: Security Robustness and Risk Profiles}

After resolving semantic inconsistencies, we focus now on security. GenAI models can introduce vulnerabilities into software projects in several ways: 1) Replication of common insecure patterns present in their vast training datasets (e.g., inheriting vulnerabilities commonly found in public repositories like GitHub); 2) Inability to anticipate novel attack vectors or malicious use cases because their competence is fundamentally bound by the statistical patterns learned during training---if a threat type is not well-represented in the data, the model effectively lacks the ``knowledge'' to guard against it.

Human programmers, conversely, introduce security flaws through different pathways, often stemming from their cognitive architecture and situated context: knowledge gaps (unfamiliarity with specific threats or secure coding practices), cognitive biases (such as optimism bias, underestimating risks), attentional limitations leading to oversight, or external pressures (e.g., prioritizing feature delivery over security hardening due to client or management demands). Crucially, humans can intentionally deceive---deliberately writing malicious code. This was highlighted by~\cite{10.1145/358198.358210} in their ACM Award Lecture, reflected in the abstract:

\begin{quote}
    To what extent should one trust a statement that a program is free of Trojan horses? Perhaps it is more important to trust the people who wrote the software.
\end{quote}

However, from a functionalist perspective, intentionally malicious code is not strictly a \textit{cognitive error} in the sense of a processing failure; rather, it represents a deliberate misalignment of goals, leveraging cognitive capabilities for harmful ends. Lacking intentional states like human malice, an AI model unintentionally generates vulnerable code as a side-effect of its statistical pattern-matching architecture operating on potentially flawed data.

However, a model's architecture and training data are not immune to human intent. Following Thompson's reflection in the context of GenAI, malicious actors could deliberately \textit{poison the training data} or fine-tune a model to consistently generate vulnerable code---regardless of the prompt, programming language, or the user's good intentions. Detecting or preventing intentionally compromised models remains a major challenge. Mitigation likely relies on \textit{pragmatic verification methods} applied at the system level: rigorous auditing of training data, model behavior testing (including adversarial testing), and potentially community-driven validation or consensus protocols. Evaluating suspect models against trusted, widely-used benchmark models could serve as a form of \textit{distributed epistemic validation}, akin to a network of \textit{cooperating epistemic agents} (though engineered and overseen by humans), applying practical checks for anomalous or harmful outputs~\citep{Simon2015}.

Therefore, vulnerabilities from AI and human programmers differ fundamentally in their underlying causes. AI-generated vulnerabilities typically arise as unintended consequences of the model's statistical learning process operating on its data corpus; they lack inherent intentionality. The only ``intention'' involved might be an external human one manipulating the training environment. Human programmers, on the other hand, can produce vulnerabilities both unintentionally due to cognitive limitations and processing errors, and intentionally, driven by specific goals to deceive or cause harm. This distinction underscores the different \textit{risk profiles} associated with each source and necessitates distinct pragmatic strategies for mitigation, verification, and trust calibration.

A pragmatic approach to addressing security errors in AI architectures involves adopting the \textit{intentional stance}, as proposed by~\cite{Dennett1971-DENIS}. This does not imply anthropomorphizing the model---we know that a GenAI system does not possess human-like beliefs or desires---but rather accepting this attribution for its explanatory utility: it helps us predict and reason about the system’s behavior. As said Dennett: ``The decision to adopt the strategy is pragmatic, and is not intrinsically right or wrong.''~\citep[p. 5]{Dennett1971-DENIS} This approach seeks cooperation. As we interact more with the model in human–machine cooperation contexts, we refine our intentional stance, improving both understanding and control of the system.

\subsection{Dimension 3: Epistemic limits}

\subsubsection{Context Integration}
Since semantic coherence and security robustness are concerns that affect both GenAI and humans, epistemic limits also pose a challenge. AI models like LLMs operate with a defined \textit{context window}, representing the maximum amount of textual information (tokens) they can process in a single input-output exchange. While larger context windows theoretically allow models to handle more external information, their practical ability to effectively utilize this entire window and maintain coherence across it can be limited. Empirical evidence~\citep{hosseini2024efficientsolutionsintriguingfailure} suggests performance often degrades as the context window fills, making information retrieval unstable or inaccurate, particularly for details far from the immediate focus.

This architectural constraint causes characteristic failure modes in large software projects. The model may struggle to integrate information beyond the immediate prompt or context window, potentially ignoring project-wide coding conventions, architectural patterns, or implicit requirements conveyed through documentation or discussions outside the provided context. This can result in code generation that is \textit{locally plausible} (correct within a small snippet) but \textit{globally incoherent} or incorrect when considered within the larger system architecture or project standards. This reflects a fundamental limitation in the model's shallow, text-bound context processing mechanism.

Human programmers face analogous---though mechanistically different---challenges in managing context within large, complex systems. Their failures can stem from limitations in retrieving relevant knowledge from long-term memory, communication breakdowns within teams leading to inaccurate or incomplete information transfer, or the inherent difficulty in maintaining accurate and complete \textit{mental models} of vast, intricate software systems, especially those involving opaque components or legacy code. Human cognitive architecture, while capable of integrating broad, multi-modal information streams (text, diagrams, conversations, past experiences) and forming rich situational models, is nonetheless fallible due to constraints on attention, working memory, and the reconstructive nature of long-term memory.

However, not all programmers ``battle'' with code identically; experience matters. With experience, programmers develop rule-based intuition, enabling them to sense when ``something is wrong'' just by looking at code. Interestingly, a study by~\citep{Kuo2024} found that less experienced programmers are more likely to deem syntactically correct code unacceptable if its semantic relationships are implausible. In contrast, experts often downplayed these critical aspects of semantic incongruity. This demonstrates how experience shapes code comprehension, leading programmers at different levels to potentially derive different understandings. Could a similar dilution of ``comprehension'' occur when an LLM is trained on limited data?

Thus, both AI and human systems show functional limitations in context integration, though these arise from vastly different underlying architectures. The AI's limitation is primarily defined by its text-bound processing and fixed context window size/effectiveness, whereas human limitations stem from the constraints of biological cognitive mechanisms operating within complex informational and social environments. This architectural gap yields distinct error patterns in large-scale coding, requiring tailored strategies (e.g., prompt design for AI, documentation for humans, and modularity for both) to manage context-related risks.

 \subsubsection{Adaptability and Generalization Boundaries}\label{subsec:adaptability_generalization_boundaries}

GenAI models often perform poorly on tasks or data far outside their training distribution. Floridi's Levels of Abstraction (LoA)~\citep{Floridi2008} help clarify this limitation. At a behavioral LoA focused on generating code snippets, a model might appear competent, seemingly ``understanding'' the task. However, if we shift to a LoA that demands deep semantic reasoning, causal understanding, or genuine novelty, the model's performance degrades. Such expectations may exceed the design of current architectures. What emerges instead is the model's true mechanism: sophisticated statistical pattern matching, not human-like comprehension. The failure lies not in applying the LoA, but in the mismatch between the AI's architectural capabilities and the higher-level expectations of certain LoAs. For example a model may perform well at the LoA of code generation but fail at factual accuracy, often \textit{hallucinating} plausible yet false information.

This architectural limitation contrasts with human adaptability, though humans also have limits. Learning novel coding strategies (e.g., switching computational models) demands cognitive restructuring and may challenge ingrained habits. Expert programmers understand that programming languages such as Python and Haskell, or Java and Prolog, or even JavaScript and Rust, differ significantly in their fundamental approaches (computational models) and grammatical rules (syntax); it requires a different way of thinking about computational artifacts. Human adaptability is bounded by individual limits in talent, prior knowledge, and cognitive flexibility. However, while AI models mainly interpolate within their training distribution, human cognition enables more flexible abstraction, analogical reasoning, and causal intervention---allowing broader, though still imperfect, generalization and adaptation to novelty.

Both human programmers and AI models thus operate within distinct \textit{operational boundaries of competence}. The AI's boundaries are sharply defined by its training data and architectural constraints (e.g., reliance on statistical correlations, limitations in extrapolating beyond learned patterns). These limits become evident when the AI is tasked with handling genuinely novel abstractions or problem types not well-represented in its training data. Human boundaries, while also real, stem from different cognitive and experiential factors and allow for different, often broader, forms of generalization and adaptation, albeit imperfectly. Understanding these different boundaries is crucial for deploying AI effectively and recognizing tasks where human adaptability remains superior. This suggests that this dimension (or indeed, others) might be associated with one or more LoAs. 

However, the utility of Floridi's LoA extends beyond serving merely as an analytical tool to describe and explain behavioral differences or epistemic limitations when observing a system (such as a GenAI model) from perspectives focused on different sets of observables. More significantly, for the present analysis, LoAs become particularly insightful when applied to understanding the \textit{interactions} between the four dimensions of ``Architectures of Error'' themselves (as discussed in Section~\ref{subsec:interaction}).

\subsection{Dimension 4: Control Mechanisms and Debugging}\label{subsec:dim4}

\subsubsection{Output Consistency}

Architectural differences fundamentally shape how AI and human systems handle informational limits (Dimension 3), and their distinct control mechanisms, along with the consistency of their outputs, pose additional epistemic challenges related to reliability and predictability.

GenAI models exhibit inherent \textit{stochasticity}, often producing variable code outputs even for identical inputs and low temperature\footnote{Temperature is a parameter controlling randomness in LLM output; lower values yield more focused, less varied (though often still non-deterministic) responses.} settings. This non-determinism by design can lead to high sensitivity, where minor prompt variations trigger unpredictable changes in the generated code.

Human programmers, driven by distinct cognitive architectures, also produce variable outputs from different sources. Human variability often stems from structured factors like \textit{learning} and \textit{experience}, strategic exploration of solutions, cognitive state (e.g., fatigue impacting performance), or biases influencing choices among valid options. Regarding cognitive biases, a programmer may consistently favor one programming language, assuming its universal suitability. This bias can skew design decisions, producing suboptimal computational artifacts that conflict with best software design practices or community consensus. Unlike the stochastic variability of GenAI, human errors from such biases are more structured, shaped by cognitive control, experience, and learned preferences---even when they diverge from prevailing epistemic norms in the technical community.

This output variability impacts both biological and artificial systems across key pragmatic dimensions: 

\begin{enumerate}
    \item \textit{Replicability:} Human programmers exhibit partial replicability. Asked to solve the same bug again, a programmer might retrieve their previous solution, perhaps with minor implementation variations (e.g., different variable names, slight abstraction changes). Significant divergence usually requires new learning or insight; otherwise, they might converge on their initially favored approach. AI models, however, due to their probabilistic nature, might consistently produce slightly different variations or potentially remain stuck generating suboptimal or flawed code patterns derived from their training data, requiring architectural changes or retraining with new data for fundamental improvement.

    \item \textit{Debugging:} Output variability inherently complicates debugging for any system. Pragmatic strategies like external verification systems, comprehensive logging, and static/dynamic code analysis tools become essential for managing systems with non-deterministic elements~\citep{10.1145/2927299.2940294}. However, debugging GenAI outputs faces unique hurdles stemming from the interaction between its variability and its specific limitations in semantic coherence (Dimension 1) and context handling (Dimension 3), making error tracing more challenging than for typical human-generated variations.

    \item \textit{Reliability Engineering:} Strategies for ensuring system reliability must account for the \textit{source} of output inconsistency. While perfect consistency enhances trust, neither GenAI nor human programmers guarantee it. Humans, however, possess a distinct advantage through \textit{cooperative feedback loops}: they can solicit targeted feedback, clarification, or corrections from other agents (e.g., more experienced colleagues) who possess relevant causal understanding or external knowledge. Current AI models, even with sophisticated training or internal critique mechanisms, are largely constrained by their probabilistic architecture; their response to feedback or self-correction prompts is still fundamentally shaped by statistical patterns rather than an ability to integrate targeted, corrective knowledge from an external source in the same way a human can engage in reasoned dialogue. This architectural limitation impacts the effectiveness and efficiency of achieving high reliability.
\end{enumerate}

\subsubsection{Process Traceability}\label{subsubsec:process_traceability}

Part of a broader internet culture movement, ``Vibe Coding'' (a term recently introduced by Andrej Karpathy to describe programmers guiding AI models instead of writing code directly) highlight the challenge posed by the \textit{opacity} of LLMs. The internal state of these models, often represented by high-dimensional vectors derived from statistical correlations, makes it exceedingly difficult to reconstruct the specific processing pathway (the ``correlational chain'') that led to a faulty output. This lack of process traceability exacerbates the debugging difficulties introduced by output inconsistency.

Although human introspection is fallible~\citep{Caporuscio2021}, humans typically retain some ability to access or reconstruct parts of their problem-solving process. They can often identify specific knowledge gaps or flawed assumptions that contributed to an error. For instance, a programmer might recognize their inability to build a game engine stems from insufficient knowledge of physics or trigonometry. Current GenAI models lack this metacognitive ability to assess their own knowledge limitations or explicitly signal ``I don't know.'' Their probabilistic architecture drives them to produce the most likely output, lacking autonomous means to assess quality; meaningful refinement typically depends on external human feedback.

This difference in internal representation is crucial. AI systems rely on distributed probabilistic vectors. Human reasoning, in contrast, is often more symbolic, step-wise, and causally structured. This divergence leads to distinct challenges. It affects not only how we trace program execution. It also complicates how we \textit{diagnose the nature of the required debugging}. Not all debugging is equivalent; different failure modes necessitate different diagnostic strategies. An AI might repeatedly generate plausible but incorrect solutions due to context limitations or by overweighting common but flawed patterns from its training data (effectively getting stuck in a local probability maximum). A human programmer uses experience and mental simulation to reason beyond immediate statistical cues, navigating solutions conceptually rather than linearly. This difference shapes the most effective debugging tools and cognitive strategies for each system.

When attempting to ensure the traceability of results in software development involving GenAI, a human programmer should adopt a stance informed by Rescher's conception of rationality~(\citeyear{Rescher1992-RESRAP-5}). Rescher argues:

\begin{quote}
 Rationally is to make use of one's intelligence to figure out the best thing to do in the circumstances. Rationality is a matter of deliberately doing the best one can with the means at one's disposal---of striving for the best results that one can expect to achieve within the range of one's resources---specifically including one's intellectual resources. Optimization in what one thinks, does, and values is the crux of rationality.
\end{quote}

Rationality here means wisely choosing the best action given the circumstances. For software projects using LLM-generated code, this requires ongoing pragmatic evaluation of:

\begin{itemize} 
    \item The \textbf{means} employed, including:
    \begin{itemize}[label=\textendash, leftmargin=*] 
        \item available resources (computational, financial, temporal);
        \item the skills and expertise of the programming team;
        \item chosen development methodologies and V\&V processes;
        \item the specificity and clarity of prompts provided to the LLM;
        \item the capabilities and known limitations of the LLMs utilized;
        \item and the overall traceability of the AI-assisted workflow.
    \end{itemize}
    \item The \textbf{ends} pursued, specifically:
    \begin{itemize}[label=\textendash, leftmargin=*] 
        \item the human-defined objectives for the software;
        \item the values embedded in or guiding the project (e.g., reliability, security, maintainability, fairness);
        \item and adherence to functional and non-functional requirements.
    \end{itemize}
\end{itemize}
This evaluation must weigh real-world constraints---budget, deadlines, ethics. Rescherian rationality provides a key framework for responsibly managing GenAI's opacity and traceability challenges. From a Rescherian perspective, the ultimate goal is the pragmatic success of the software endeavor: delivering functional, reliable, and secure systems that fulfill defined objectives within real-world constraints. Recognizing ``Architectures of Error''  is a starting point for addressing evolving technology (the \textit{means}) and shifting project goals (the \textit{ends}). The key question, then, is how an evolving understanding of these architectures—including emerging failure modes—can help align changing means and ends to secure pragmatic success.

\subsection{Dimensions interactions}\label{subsec:interaction}

A crucial further question arises: are these dimensions isolated phenomena, or do they interact? I argue that they are not isolated; indeed, the complexity of the architectural divergence is compounded by the interrelations between these dimensions, with some interactions being more evident or impactful than others. These potential interactions are summarized in Table~\ref{tab:dimension_interactions}. 

I argue that one way to elucidate the interaction between the dimensions of ``Architectures of Error'' is by analyzing them through their associated Levels of Abstraction (LoAs). For this purpose, as introduced in Subsection~\ref{subsec:adaptability_generalization_boundaries}, I will employ Floridi's Levels of Abstraction~\citeyearpar{Floridi2008} as a conceptual tool.

Let each dimension $D_i$ (where $i \in \{1, 2, 3, 4\}$) encompass multiple associated error types, $e \in D_i$. Each error $e$ can be characterized by, or analyzed at, one or more LoAs. Interactions between dimensions $D_i$ and $D_j$ (where $i \neq j$) can then be understood in terms of the relationships between the LoAs associated with their respective errors, $e_{D_i}$ and $e_{D_j}$. A compelling interaction, leading to a deeper systemic understanding, often arises when errors from different dimensions are analyzed at \textit{different yet relatable} LoAs. Conversely, if errors from distinct dimensions are only considered at very similar or identical LoAs, their interaction might appear weak or less insightful for understanding systemic failure.

Consider the interaction between $D_1$ and $D_3$ in a GenAI system (see first row in Table~\ref{tab:dimension_interactions}), analyzed through distinct LoAs. A higher LoA captures global semantic coherence—meaning, logic, and project-wide constraints---while a lower, architectural LoA reflects internal processing limits, such as context window size. A failure at the lower LoA (e.g., exceeding the context window) can propagate upward, causing semantic incoherence at the higher LoA.

This interplay shows how architectural constraints ($D_3$) can drive functional failures ($D_1$), and highlights the value of LoAs for tracing errors across abstraction layers. Such multi-level analysis sharpens our understanding of GenAI limitations and the structural relations between error types.

These observable interactions provide a basis for extending the framework as technology evolves---introducing new errors and, consequently, new interactions. By mediating them through LoAs, these future interactions can be explained pragmatically.

\rowcolors{2}{rowalt}{lavenderblush} 
\setlength{\tabcolsep}{8pt}
\renewcommand{\arraystretch}{1.3}

\begin{longtable}{|>{\raggedright\arraybackslash}p{2.3cm}|>{\raggedright\arraybackslash}p{5cm}|>{\raggedright\arraybackslash}p{5cm}|}
\caption{Interactions and Compounding Effects in GenAI Code Error Dimensions (D1 = Semantic Coherence, D2 = Security, D3 = Epistemic Limits, D4 = Control / Consistency).}
\label{tab:dimension_interactions} \\
\hline
\rowcolor{headergray}
\textbf{Interacting Dimensions} & \textbf{Description of Interaction and Compounding Effect} & \textbf{Illustrative Example} \\
\hline
\endfirsthead
\hline
\rowcolor{headergray}
\textbf{Interacting Dimensions} & \textbf{Description of Interaction and Compounding Effect} & \textbf{Illustrative Example} \\
\hline
\endhead

\hline
\endfoot

\hline
\endlastfoot

\textbf{D1 and D3} &
Limited context window (D3) impairs global semantic coherence (D1), causing locally plausible but globally flawed code. &
Model ``forgets'' early context (D3), generating code with logic (D1) inconsistent with project system. \\

\textbf{D1 and D2} &
Poor semantic understanding (D1) leads to misuse of security primitives or flawed logic, creating vulnerabilities (D2) despite syntactic correctness. &
Correct syntax for locks but flawed semantic implementation (D1) causes race conditions (D2). See Appendix~\ref{app:atomictask}. \\

\textbf{D1 and D4} &
Model opacity (D4) hinders diagnosing semantic errors (D1); stochasticity (D4) causes unpredictable appearance/disappearance of these flaws. &
A function yields nonsensical output intermittently (D1 flaw) due to untraceable model ``reasoning'' (D4). \\

\textbf{D3 and D2} &
Training data limitations (D3) cause blindness to novel attack vectors (D2); small context windows (D3) miss project-wide security needs (D2). &
No defense against zero-day exploits (D2) due to training data gaps (D3); or, locally secure code is systemically insecure due to narrow context (D3). \\

\textbf{D3 and D4} &
Near epistemic limits (D3), GenAI output becomes erratic and inconsistent (D4), complicating debugging of boundary-case failures. &
Generating code for a novel algorithm (D3) yields highly variable and incorrect outputs (D4), even at low ``temperature''. \\

\textbf{D2 and D4} &
Non-determinism (D4) leads to intermittent security vulnerabilities (D2); opacity (D4) hinders auditing for subtle, AI-specific flaws. &
SQL injection vulnerability (D2) appears sporadically due to model stochasticity (D4), evading simple tests. \\

\textbf{General Compounding Effect:} \footnotesize\textbf{D1+D2+D3+D4} &
The interaction among these dimensions reveals a compounded epistemic fragility: limited context integration (D3) gives rise to semantic distortions (D1) and increases the system’s vulnerability to security failures (D2), while the stochastic and opaque nature of the model’s architecture (D4) renders such failures difficult to trace, reproduce, or even fully explain. &
Misinterpretation of global state (D3 error) leads to logic bugs (D1 and D2), which are inconsistent and hard to trace (D4), baffling developers. \\

\end{longtable}

\section{Functional Consequences for Software Engineering Practice}\label{sec:implications}

Philosophical reflection on software engineering should ultimately aim to influence practice. In this spirit, a central argument of this paper is that understanding the distinct architectural origin and the failure modes previously analyzed (Section~\ref{sec:comparative_analysis}) in relation to code generation has direct and practical implications for development, testing, and management of software involving GenAI. 

Consider the traditional approach to learning programming, described by Carlos Baquero in ``The Last Solo Programmers'' before the widespread adoption of GenAI tools:
\begin{quote}
Until quite recently, programmers acquired their skills by (ideally) starting with a mathematical and abstract problem-solving background and practicing programming. First, they would write simple programs and then gradually move on to more complex ones, learning how to break down and compartmentalize larger problems along the way. Traditionally, this was done in plain text editors such as VI and EMACS.~\citep{baquero2025cacm}
\end{quote}

This traditional, incremental learning---focused on building cognitive models and debugging reasoning---contrasts sharply with the shifting interaction model of GenAI. While programming tools have evolved significantly (e.g., rich IDEs), these changes are minor compared to the fundamental architectural differences between human and AI code generation (Section~\ref{sec:comparative_analysis}).

Against this backdrop, GenAI introduces new layers of complexity and opportunity. Recognizing the distinct architectural strengths and weaknesses of both human cognition and GenAI systems is crucial for effectively integrating these tools. In the following subsections, I identify four key functional implications for software engineering practice, focusing on how to leverage GenAI's capabilities while mitigating its inherent risks through an understanding of these foundational differences.

\subsection{Philosophical Challenges in Verifying and Validating GenAI-Generated Code}\label{subsec:v_v}

Standard software engineering techniques distinguish between verification (``Are we building the computer program correctly?'') and validation (``Are we building the right computer program?''). While essential, these traditional V\&V approaches are often insufficient for code generated by GenAI due to its specific failure modes rooted in its distinct architecture.

For instance, unbalanced or low-quality training data can lead LLMs to generate biased code, reflecting societal biases related to age, socioeconomic status, political affiliation, race, sexual orientation, etc. Early LLM versions might have attempted to generate simplistic, incorrect, and biased functions in response to prompts like ``Provide a Python function to determine if a person is poor''~\citep{huang2024biastestingmitigation}. While current state-of-the-art models (e.g., GPT-4o, Gemini-2.5-Pro, Claude-4) are often trained to recognize and refuse such reductive or underspecified prompts concerning complex, multifactorial traits, the underlying potential for bias rooted in data persists as a significant challenge inherent to the statistical learning paradigm.

The distinct error profiles of GenAI, particularly its susceptibility to semantic incoherence (Dimension 1), challenge traditional V\&V approaches centered on syntactic checks and unit tests. Consequently, ensuring GenAI-generated code reliability raises the philosophical question of adequate \textit{epistemic warrant}, suggesting a need for more rigorous semantic testing and potentially statistical methods to address computational artifact fallibility. Validation must specifically target the failure modes characteristic of GenAI, such as susceptibility to biased data patterns and difficulties with complex semantic coherence, rather than assuming failure modes identical to typical human programming errors.

Similarly, the unique security risk profiles of GenAI (Dimension 2), such as the replication of insecure patterns or blindness to novel attack vectors, prompt a philosophical re-evaluation of security assurance. This involves considering the epistemic grounds for trusting code whose vulnerabilities may stem from statistical learning rather than intentional design flaws, suggesting that standard vulnerability checks may be insufficient without targeted, architecture-aware auditing.

 V\&V must also address epistemic limits (Dimension 3) by explicitly ensuring global coherence and compliance with project-wide constraints, which often exceed the effective context window of GenAI models. Similarly, testing strategies for non-deterministic outputs (Dimension 4)---such as multiple execution runs, statistical significance analysis of outputs, and robust regression testing given prompt sensitivity---are crucial. This remains critical, as even with temperature $0$, GenAI models offer increased stability---not determinism~\citep{10.1145/3697010}.

Adopting advanced V\&V methods is key to ensuring reliable and effective software development with GenAI tools. These methods are pragmatically warranted due to their ability to address the unique failure modes inherent in AI architectures. While this approach cannot guarantee the lack of errors, it does allow for the measurement of failure rates---an essential step toward mitigating them. As Rescher noted regarding human life, ``Evaluation is an essential requisite of human existence.''~\citep[ p.~3]{Rescher2017-go}; we could paraphrase this in the context of AI: ``Evaluation is no less essential to the existence of code generated by a GenAI model.''

Together, these dimensions underscore a key epistemic distinction: verifying code generated by GenAI models is not equivalent to verifying other types of LLM-generated data, such as natural language text, summaries, or question-answering outputs. This distinction arises because programming languages are formal systems. Consequently, bridging the gap between an LLM's stochastic architecture and the deterministic requirements of a formal language necessitates a distinct approach to correctness and verification. This challenge, however, is not entirely novel within computer science. 

In the field of computational optimization, for instance, both exact and stochastic algorithms are employed. Hybridizing these approaches---combining exact and stochastic components---demands rigorous statistical analysis to ascertain the quality of solutions, particularly since the stochastic element ensures that multiple executions of the same hybrid algorithm may not yield identical results.\footnote{An area that inherently engages with hybridization is Metaheuristics. Rather than a single algorithm, a metaheuristic is a general framework for working with various heuristic methods, especially in the context of optimization problems. These approaches can often be combined with exact methods---procedures that guarantee finding the optimal solution for a given problem instance, such as Integer Linear Programming~\citep{10.5555/3279208}.

Their epistemic challenges echo those of GenAI: How can algorithms that rely on random operators achieve strong empirical performance without a theoretical account of their behavior?} Tests are beginning to serve not only as a mechanism to assess the correctness and quality of an algorithm, but also its robustness; they become an epistemic warrant that allows us to work with an opaque architecture (GenAI models). This is fundamental concern in any such hybridization. Integrating AI-generated code (from a stochastic architecture) with human-authored code (which typically aims for deterministic logic) into larger software systems presents a similar challenge. To account for these inherent architectural differences, development teams must therefore craft bespoke methodologies for integration, correctness assurance, and robustness validation.

Recently, AlphaEvolve~\citep{Novikov_undated-wj}, a system of LLM-based agents developed by Google, has introduced generative techniques to improve algorithms through an evolutionary process, modifying code by replacing segments with newly generated ones. Rather than altering the architecture of GenAI models, this approach adds a formal layer, resulting in a hybrid system. To ensure the reliability of this hybridization, formal validation methods have been integrated.

This gives rise to new epistemic concerns: How can we trust that code produced by a stochastic architecture---with its well-known limitations and propensity for error---can lead to genuinely novel and practically useful solutions?

\subsection{The Attribution Problem: Rethinking ``Responsibility''}

Discussions of ``responsibility'' concerning AI often focus on the ethical and social implications of its deployment. However, when applied to GenAI-driven code generation, simply attributing responsibility to the AI itself is an oversimplification. Instead, responsibility---including that related to code generation---must be understood systemically and pragmatically within the context of human-AI collaborative development, where code generated by GenAI models coexists and interacts with human-written code. In this scenario, responsibility becomes distributed across these two distinct architectural sources.

This \textit{human-AI symbiosis} in code generation raises critical questions regarding accountability:

\begin{enumerate}[label=(\arabic*)]
    \item How can we safely integrate code from a GenAI system---a system that is neither fully reliable nor completely verifiable in its internal processes---into legacy or new software systems?
    \item If an integrated human-AI codebase malfunctions, who, or what, bears responsibility for the failure?
\end{enumerate}

The AI’s inherent lack of intention, understanding, and autonomous self-correction (as discussed in Section~\ref{sec:comparative_analysis}) makes assigning human-like responsibility to the model problematic.  Adopting a pragmatic approach to responsibility offers a more constructive path. The coexistence of human and AI-generated code necessitates clearly defined roles, robust processes, and explicit accountability structures within the human teams utilizing these AI tools. Key questions then arise: Who is responsible for defining and enforcing V\&V standards for AI-generated components? Who ultimately approves the integrated code before deployment? The focus moves to risk management and ensuring procedural safeguards.

Regarding the first question (1), a pragmatic perspective dictates that the evaluation of AI-generated code must be tailored to its specific ``Architecture of Error.'' It cannot be assessed using the same assumptions or methods applied to human-written code. Just as their origins differ, so too must their evaluation methodologies. This, in turn, raises further practical questions, such as how to effectively integrate testing regimes for AI-generated code with those for code developed by human teams.

Concerning the second question (2), consider a security vulnerability (Dimension 2) arising from AI-generated code. If poisoned data led to this vulnerability, is the data provider, the model trainer, or the deploying engineer who failed to implement sufficient verification measures responsible? The opacity of GenAI models (Dimension 4: Control Mechanism and Debugging) makes it difficult to understand precisely why a model failed. Therefore, maintaining clear traceability of how AI tools are used within the development team becomes indispensable. This includes meticulous record-keeping of the experimental methodology, prompts used, outputs obtained, model versions selected, and parameters chosen, providing at least a partial audit trail for any integration of cognitive and artificial architectures. If this is not addressed, could the human–AI symbiosis in code generation entail that programmers bear a new ``duty of care'' when using AI tools? And in the event of failure, does responsibility fall entirely on the programmer, or is it mitigated by the involvement of AI systems?

\subsection{Trust Calibration Based on Mechanism Awareness}

 Discourse surrounding GenAI models for code generation often polarizes into two extremes: uncritical over-trust in the technology's capabilities or excessive skepticism regarding its utility. A philosophically grounded approach to human-AI collaboration in code generation suggests that trust must be epistemically calibrated. This calibration should be informed by a nuanced understanding of the distinct architectural strengths and weaknesses of both AI and human cognition, as outlined by their respective ``Architectures of Error''. Such mechanism-aware trust calibration is a pragmatic necessity for productive human-AI collaboration.
 
This calibration can draw directly on insights from each analyzed dimension:

\begin{itemize}
    \item \textbf{Semantic Coherence (Dimension 1):} While the syntactic generative capabilities of AI models may warrant a degree of instrumental trust, their susceptibility to hallucinations and reliance on statistical patterns---rather than deep, model-based understanding---presents a significant epistemological challenge. This necessitates a critical stance towards the semantic coherence and logical integrity of generated code, demanding rigorous verification practices that explicitly account for these underlying architectural limitations.
    
    \item \textbf{Security Robustness (Dimension 2):} Exercise caution regarding security, pragmatically assuming the potential for replicated insecure patterns from training data or an inability to anticipate novel threats. Verification should target these specific AI-driven risks.
    \item \textbf{Epistemic Limits (Dimension 3):} Limit trust for tasks demanding broad contextual integration beyond the model's effective context window or requiring robust generalization to out-of-distribution scenarios where its pattern-matching capabilities falter.
    \item \textbf{Control Mechanisms and Debugging (Dimension 4):} Methodologies must account for inherent output inconsistency and stochastic variability in AI-generated code. Development processes should proactively plan for this non-determinism in testing, debugging, and integration strategies.
\end{itemize}

Trust calibration, in this sense, is a core \textit{pragmatic epistemic practice}, as Rescher might articulate. We continuously adjust our reliance on a method or tool based on accumulating evidence of its effectiveness, reliability, and characteristic failure modes. A fundamental understanding of the underlying mechanisms of both biological and artificial architectures---their distinct ``Architectures of Error''---is therefore indispensable for achieving this effective calibration and fostering a more robust and reliable software engineering practice involving GenAI.

\subsection{Human Cognitive Adaptation and System Effects}\label{subsec:human_adaptation}

Recent studies show that integrating GenAI tools like ChatGPT and GitHub Copilot is fundamentally transforming software development and its human-computer interaction paradigms~\citep{Ulfsnes2024}. For instance, platforms like Stack Overflow have reportedly experienced decreased traffic, possibly because programmers increasingly rely on LLMs for solutions. Interestingly, despite LLM-generated code sometimes containing more errors than Stack Overflow answers, some programmers prefer LLM outputs due to their perceived comprehensiveness and well-articulated presentation~\citep{10.1145/3613904.3642596}.

This evolving dynamic presents a dual-edged scenario. On the one hand, GenAI tools can offer positive effects, such as enabling programmers to work more efficiently, potentially accelerating learning processes, and increasing motivation by alleviating tedious and repetitive tasks. There are even indications of shifts in teamwork dynamics, with software engineers turning to GenAI for assistance instead of consulting co-workers, thereby impacting traditional learning loops within agile teams. On the other hand, significant concerns arise. Studies highlight a tendency towards \textit{over-reliance} on LLM-generated code, particularly among less experienced programmers, who may accept suggestions without sufficient critical scrutiny. This creates a paradox: AI tools intended to augment human productivity instead demand substantially more human validation effort due to the distinct ``Architectures of Error'' explored in this paper.

These positive and negative aspects raise important questions about human cognitive adaptation and the broader systemic impact of integrating GenAI into software engineering:

\begin{itemize}
    \item \textbf{Change in Core Skills:} A decreased focus on writing boilerplate code (where GenAI often excels) may be accompanied by an increased demand for skills in prompt engineering, critical evaluation of AI-generated outputs, debugging complex and potentially opaque errors (implicating Dimensions 1 and 4), high-level architectural design, and effective context provision for AI (Dimension 3).
    \item \textbf{Deskilling versus Reskilling Dilemma:} There is a potential for the atrophy of fundamental coding skills if over-reliance becomes widespread, contrasted with the opportunity for developing new, valuable human-AI interaction and co-creation competencies.
    \item \textbf{Impact on Cognitive Biases:} Interaction with AI systems that exhibit superficial competence (Dimension 1) could exacerbate existing human cognitive biases, such as automation bias or confirmation bias. Furthermore, the output inconsistency of AI (Dimension 4) might introduce new workflow disruptions or cognitive load challenges.
    \item \textbf{Evolving Team Dynamics and Practices:} The integration of AI-generated code necessitates rethinking communication protocols within teams (related to Dimension 3). For example, how will peer review processes adapt to code with mixed authorship (human and AI)? Could GenAI models themselves be incorporated into the review process, and what would be the implications of AI reviewing AI-generated code?
\end{itemize}

Integrating two distinct software programs usually requires some adaptation. This challenge is amplified when interfacing systems with fundamentally different architectures, such as human-written code and GenAI-generated code. Unlike adapting two human-authored software components, where discrepancies might be resolved through dialogue and mutual adjustment leading to an agreed-upon interface, such direct negotiation is impossible when one party delivering a code artifact is non-human. We cannot demand the same kind of accountability from a GenAI model as we would from a human programmer, precisely because it lacks human-like consciousness, intentionality, or understanding of downstream consequences.

However, this absence of inherent AI accountability can be pragmatically managed by clearly assigning responsibility to the humans overseeing the AI-generated code. While these human agents may not bear full responsibility for an unforeseen architectural failure within the model itself, they can and must assume responsibility for the \textit{pragmatic processes following code generation}. This includes: integrating the AI-generated component into the human-developed system; conducting rigorous validation and verification tailored to the GenAI model's specific ``Architecture of Error''; and defining the criticality and potential impact of the AI-generated code within the larger software system. Thus, responsibility shifts from the non-comprehending AI to the human-led governance and integration framework overseeing its output.

\section{Discussion}\label{sec:discussion_broader_theories}

This section situates the ``Architecture of Error'' framework within two perspectives: a sociotechnical view of AI errors and a normative approach from eXplainable AI. It concludes by addressing likely objections to the framework.

\subsection{Integrating Architectural Insights with Sociotechnical Theories of AI Error}

\citet{Barassi2024Toward} offers a contrasting theory of AI error, focusing on its sociocultural, political, and economic dimensions, and treating error as a sociotechnical construct shaped by broader social dynamics. Her ``theory of AI errors'' aims to explain how external, contextual factors shape the production of erroneous knowledge by AI systems. In contrast, my approach focuses on internal architectural differences between AI and human cognition as the source of distinct failure modes, emphasizing their epistemological and pragmatic implications for interacting with and verifying generated code. This work offers a foundational, proximate explanation for these errors (Section~\ref{sec:comparative_analysis}), while Barassi addresses broader sociocultural and production contexts. Both perspectives are arguably essential for a full understanding.

Grasping the ``Architectures of Error'' at a functional and epistemological level (my focus) can be seen as a prerequisite for effectively addressing the sociocultural implications and governance strategies highlighted by Barassi (or future theories of AI errors in sociotechnical domain). Understanding \textit{how} and \textit{why} these systems fail technically is crucial for assessing their societal impacts and designing appropriate regulations. It is in this sense that adopting a mechanistic (Dennett) and methodological-pragmatic (Rescher) stance offers distinct utility: it provides an applicable framework for enhancing software engineering practices, refining Verification and Validation processes, clarifying functional responsibility attribution, and improving trust calibration at the level of development and direct tool interaction (see Section~\ref{sec:implications}).

\subsection{Do Normative Frameworks for XAI Adequately Address Systems with Inherent Stochasticity?}

A substantial portion of research in eXplainable AI (XAI) is dedicated to developing normative frameworks that address how, what, and why explanations should be provided. A notable example is Zednik’s influential work, \emph{Solving the Black Box Problem: A Normative Framework for Explainable Artificial Intelligence}~\citep{Zednik2021}. These frameworks play a key role in articulating the core requirements of XAI and in shaping ethical and societal discourse around AI system transparency.

A key limitation of largely normative frameworks like Zednik’s is their lack of concrete guidance for systems with complex technical specifics. This is particularly evident in models that rely on pseudo-randomness---such as the Machine Learning and Metaheuristic methods discussed in Section~\ref{subsec:v_v}. For instance, generating faithful explanations in inherently stochastic systems—such as LLMs or Metaheuristics with randomized operators—poses challenges that general normative principles alone cannot fully resolve.

This is where the "Architectures of Error" analysis provides a foundational and complementary perspective. It centers on the architectural differences and distinctive error profiles of the systems in question (see Section~\ref{sec:comparative_analysis}). These issues are especially relevant to Dimension 4 (Section~\ref{subsec:dim4}), which deals with stochasticity and opacity. The framework emphasizes that a deep understanding of these intrinsic characteristics is a necessary step before developing effective explainability or V\&V strategies (Section~\ref{subsec:v_v}).

Understanding the ``Architecture of Error'' in inherently stochastic systems is, in my view, an epistemic prerequisite for assessing which types of explanations are feasible or trustworthy. It also helps determine what pragmatic tools---such as Rescherian strategies for handling limited traceability (see Section~\ref{subsubsec:process_traceability})---are needed when full explainability is out of reach. My goal is not to replace normative XAI frameworks, but to enrich them with an epistemic and architectural foundation that addresses the technical particularities of generative models.

\subsection{Anticipated Objections}

I aim to address---though not exhaustively---four potential objections to the ``Architectures of Error'' framework defended in this paper.
\begin{enumerate}[label=\arabic*.]
    \item \textbf{Are the cognitive versus artificial architectures genuinely distinct for code generation?} One might argue that this binary architectural distinction is too stark. Humans also employ heuristics and pattern-matching (akin, in some respects, to statistical mechanisms), and emerging AI might evolve more causal and structured architectures similar to biological ones. Thus, the boundaries might not be as clear-cut as presented.
    \begin{enumerate}[label=(\alph*)]
        \item \textit{Response:} While it is true that human cognition can exhibit behaviors somewhat analogous to artificial models in certain aspects (as noted within each dimension's analysis), the \textit{nature and degree} of opacity in our own behaviors are not comparable. Consider human opacity stems from the vast complexity of biological neural networks and the inherent limits of introspection. However, this opacity exists within an architecture that also possesses more adaptive mechanisms than current AI. The opacity of current AI, rooted in its statistical-stochastic nature, presents distinct challenges that human adaptive mechanisms might circumvent or mitigate differently. Thus, even if human cognition involves processes that are not fully transparent to the individual, the underlying architecture provides different pathways for adaptation and error correction than those available to current AI. Although superficial similarities may exist at certain levels, the underlying architectural distinctions manifest clearly at different levels of abstraction.
    \end{enumerate}

    \item \textbf{Does constantly comparing GenAI's code generation capabilities with those of human programmers constitute a form of reverse anthropomorphism?}
    \begin{enumerate}[label=(\alph*)]
        \item \textit{Response:} The comparative approach seeks to establish a contrast that clearly delineates the functional and epistemic implications. It uses human cognition as a familiar, intuitively understood reference point, but without attributing human mental states to artificial architectures. The aim is analytical clarity, not an equivalence of internal states. Beyond being merely an analytical reference point, however, this comparison is a \textit{pragmatic necessity}. Given that AI-generated code must coexist and integrate with human-written code, and be utilized and maintained by human programmers, understanding the differences in their error profiles using a human standard as a baseline is crucial for effective software engineering, V\&V, and trust calibration.
    \end{enumerate}

    \item \textbf{Might the rapid evolution of AI render this framework fragile or quickly outdated?}
    \begin{enumerate}[label=(\alph*)]
        \item \textit{Response:} Although the specific manifestations of errors within the dimensions---coherence, security, epistemic limits, control mechanisms---will undoubtedly evolve with technology, these categories are likely to persist as relevant analytical lenses. Addressing such evolution necessitates a framework that can operate across different Levels of Abstraction (LoAs), such as Floridi's. This allows technological advancements (which can be seen as shifts in LoAs or capabilities within them) to be positioned and analyzed within these four fundamental dimensions, as briefly explored in the discussion on dimensional interactions (Section~\ref{subsec:interaction}).
    \end{enumerate}

    \item \textbf{Are four dimensions sufficient for this analysis?}
    \begin{enumerate}[label=(\alph*)]
        \item \textit{Response:} When proposing a systemic framework with distinct categories (dimensions, in this case), questions about their number---why not three, or ten?---are natural. I selected four because this number allows me to make and defend a pragmatically committed and coherent argument. Regardless of programming language, programmer experience, company maturity, or even the type of computation (classical or quantum), all software development efforts contend with technological limits (Dimension 3), face security challenges (Dimension 2), require control mechanisms for reliable interaction (Dimension 4), and demand coherence for utility (Dimension 1, where the pragmatic commitment becomes evident). Removing or adding a dimension would, in my view, compromise the internal coherence of this framework and its translation into practical opportunities for analysis.
    \end{enumerate}
\end{enumerate}


\section{Conclusion}\label{sec:conclusion}

Generative AI for code generation is a transformative technology that is here to stay. Its persistence is not due to any claim of infallibility; as this paper outlines, it produces inconsistencies and errors rooted in its underlying stochastic architecture. Instead, this analysis suggests GenAI currently functions best as a powerful tool augmenting human programmers in the complex task of software construction. Understanding the distinct failure modes and architectural origins of both GenAI and human cognition is thus fundamental.

To grasp the distinctions between biological and artificial architectures, I have argued for the need for conceptual clarity. The comparative analysis across four dimensions---semantic errors (statistical vs. cognitive), security flaws (pattern-based vs. oversight/intent), context limitations (window vs. mental model), and output variability (stochastic vs. structured)---unveils differences so fundamentally rooted in their respective origins that they demand ongoing critical engagement as these technologies mature.

Dennett's functionalist/mechanistic approach provides the conceptual tools to understand both ``Architectures of Error'': the artificial, thereby demystifying it and avoiding anthropomorphism; and the human, thereby recognizing its capabilities without attributing to it unfathomed ``superpowers.'' Concurrently, Rescher's framework offers a pragmatic methodology for operating effectively with both architectures despite their inherent, origin-specific errors.

Analyzing distinct ``Architectures of Error'' philosophically reveals an urgent need for further consideration---both philosophical and practical-ethical---within AI-augmented software engineering. Key among these are: the epistemic imperative for rigorous, mechanism-aware Validation and Verification; re-evaluating responsibility attribution to account for distributed agency in human-AI teams; cultivating critical evaluative capacities in programmers interacting with GenAI; and understanding the philosophical bases for trust calibration. Such considerations represent not just pragmatic necessities but deeper epistemic and ethical commitments essential for responsible software development.

Acknowledging GenAI's capabilities does not diminish the philosophical and technical challenges explored here. The central concept of distinct ``Architectures of Error'' reveals crucial symmetries and asymmetries in the operational mechanisms and failure profiles of cognitive versus artificial systems. Overlooking the specific nature of AI-generated errors risks confusion, misguided tool reliance, and flawed software development decisions. While this philosophical inquiry offers clarity for navigating the evolving GenAI ecosystem, its analyses are not definitive but rather a spur to further thought, grounded in the code itself and the human and artificial processes that create it. Ultimately, understanding these distinct ``Architectures of Error'' is a pragmatic necessity for a more reliable, responsible, and effective future in software engineering.

\section*{Declarations}
\subsection*{Availability of data and material}
No data or supplementary materials are available.

\subsection*{Competing interests}
The author declares that they have no competing interests.

\subsection*{Funding}
The author received no financial support for the research, authorship, or publication of this article.

\subsection*{Acknowledgements}
I am grateful to Raymond Turner, whose book \emph{Computational Artifacts: Towards a Philosophy of Computer Science} (2018) helped me understand that computer science can be more than merely a technical discipline.

\newpage
\appendix
\section{A prompt that stresses all facets of ``Architectures of Error''}\label{app:atomictask}

\textit{Author's note: Paradoxically, I structured this prompt with the help of an LLM; however, the fact that an LLM can craft a complex prompt does not mean it is equally competent at solving it.}

The following prompt poses a challenge even for programmer experts, as it brings together multiple software engineering concepts---such as invariants, concurrency, deep abstraction, fine-grained synchronization, multiple interacting algorithms, and semantic coherence---that are notoriously difficult to reason about when entangled. It demands the design of a complex internal architecture rather than the mere implementation of a function. Even today, such scenarios remain non-trivial for the most advanced generative models (Gemini-2.5-Pro, GPT-4o, Llama-4, Claude-4).

As prompt complexity grows, models may handle parts correctly, but global inference errors become more likely. Real-world software often involves even messier cases---integrating new code with legacy systems and entangled concepts.

\begin{lstlisting}[style=promptstyle]
Design and implement a Python class `AdaptiveHierarchicalTaskAssigner` that manages the dynamic and optimized assignment of heterogeneous task types to a pool of concurrent, specialized workers.

=====Problem Context=====

There are N unique task types, identified by integers from 1 to N. Each task type `j` has the following static and dynamic properties:

* `id_j`: The unique identifier.
* `base_cost_j`: An integer base cost associated with preparing the task.
* `value_j`: An integer value representing the reward/importance of completing the task.
* `category_j`: A string indicating its category (e.g., "IO", "CPU", "MEMORY").
* `dependencies_j`: A set of `id`s of other task types that must have been *fully assigned in previous sets* before task `j` can be considered for a new set.
* `urgency_j(t)`: A function that returns the current urgency of the task, which can change with time `t` (time is considered as a counter of "ticks" or assignment calls).

The assigner must issue "WorkPackages". Each WorkPackage should ideally contain `k_target_size` distinct task types, but this `k_target_size` is a *target*; a deviation of `+/- k_flexibility` is allowed (with `k_min = 1`).

=====Class Initialization=====

__init__(self,
initial_task_descriptors: List[Dict], # List of dictionaries, each describing a task (id, base_cost, value, category, dependencies)

k_target_size: int,

k_flexibility: int,

global_constraints: Dict, # e.g., {'max_total_sum_id': T_id, 'max_total_sum_cost': C_sum, 'min_total_sum_value': V_sum}

category_affinity_rules: Dict[str, Dict], # e.g., {'IO': {'max_per_package': 2, 'must_be_paired_with': 'CPU'}}

worker_profiles: Dict[str, Dict] # e.g., {'worker_type_A': {'specialized_categories': ['IO'], 'max_concurrent_packages': 3, 'cost_modifier': 0.8}}
)

* `initial_task_descriptors`: Defines the initial set of all tasks and their base properties.
* `k_target_size`, `k_flexibility`: Define the desired size and flexibility of WorkPackages.
* `global_constraints`: Constraints that each WorkPackage *must* satisfy (sum of IDs, sum of costs, sum of values).
* `category_affinity_rules`: Complex rules about how tasks from different categories can or must be combined.
    * `max_per_package`: Maximum number of tasks of this category in a package.
    * `must_be_paired_with`: If a task of this category is present, at least one task from the specified category must also be present.
    * Could include `cannot_be_with`: Incompatible categories in the same package.
* `worker_profiles`: Describes worker types. The assigner might use this to optimize or restrict assignments (see `assign_work_package`).

=====Core Assignment Method=====

assign_work_package(self, worker_id: str, current_time_tick: int) -> Optional[List[Dict]]:

* This method will be called concurrently by multiple workers, each identifying itself with `worker_id`.
* It must find a WorkPackage that:
1. Meets `k_target_size +/- k_flexibility`.
2. Satisfies all `global_constraints`. The sum of `base_cost_j` may be modified by the `cost_modifier` from the `worker_profile` for this calculation.
3. Fulfills all `dependencies_j` for each selected task (dependencies must have been assigned and *confirmed as completed* previously - see `confirm_completion`).
4. Respects `category_affinity_rules`.
5. Optimizes for a combined objective: Maximize `sum(value_j * urgency_j(current_time_tick))` of the package, and secondarily minimize `sum(modified_cost_j)`. The optimization strategy must be robust but not necessarily find the absolute global optimum if `N` is very large (well-justified heuristics are acceptable).
* Advanced Critical Invariants:
1. Active Assignment Uniqueness: A task ID, once part of a successfully *assigned* WorkPackage, cannot be part of any other *active* WorkPackage.
2. Temporary Reservation and Reconsideration: Tasks *considered* by a worker for a package, if that package attempt fails (e.g., doesn't meet the worker's local optimization target or a complex affinity rule), are *not* immediately permanently retired. They should return to the general pool of "available for consideration," but the same worker should be prevented from immediately retrying them in the same failed combination. Other workers, or the same worker with a different strategy, could use them. The "reservation" mechanism during package construction must be highly efficient and minimize contention.
3. Dependency Management: The state of "dependency fulfilled" is crucial and must be managed rigorously via `confirm_completion`.
* Extreme Concurrency:
    * The method must be thread-safe and designed for high concurrency. Dozens or hundreds of workers are expected.
    * Minimizing lock duration and granularity is essential. Consider using advanced concurrency primitives if appropriate (e.g., ReadWrite Locks, Software Transactional Memory concepts if they could be simulated or if the language/framework allowed, or at least concurrent data structures).
* Return Value:
    * If a valid and optimized WorkPackage is formed: a list of task dictionaries (each dict with all its relevant properties at the time of assignment, including the calculated `urgency`).
    * If not: `None`.

=====Support Methods=====

confirm_completion(self, worker_id: str, package_id: str, task_ids: Set[int], success: bool):

* Called by a worker to indicate that a WorkPackage (identified by a unique `package_id` generated during assignment) has been completed.
* If `success` is `True`, the `task_ids` are marked as "permanently completed," satisfying future dependencies. Tasks are retired from the "active" pool.
* If `success` is `False`, the `task_ids` must return to the "available for assignment" pool (possibly with an urgency penalty or a cooldown to prevent rapid, futile retries). Their "dependency fulfilled" status does not change for other tasks.

update_task_urgency_model(self, task_id: int, new_urgency_function_or_params: Any):

* Allows dynamic modification of how a task's urgency is calculated.

get_system_diagnostics(self) -> Dict:

* Returns metrics about the assigner's state: number of available tasks per category, assignment success rates, lock contention levels (if measurable), etc.

=====Additional Implicit and Explicit Challenges=====

1. Scalability with Large N: The number of task types `N` can be in the tens of thousands. Exhaustive search for combinations for optimization is infeasible. A heuristic or adaptive strategy must be designed.
2. State Complexity: The state includes not only available tasks but also assigned tasks (pending confirmation), completed tasks, and the dynamic state of dependencies.
3. Interaction of Constraints: The multiple layers of constraints (global, category, dependency, worker) and optimization objectives make finding a valid and "good" solution very difficult.
4. Handling Partial Failures: What happens if a worker takes a package but fails to process only a part of it? `confirm_completion` must handle this.
5. Adaptability: The system should be able to adapt to changes in urgency functions or even the addition/removal (infrequent) of task types at runtime (the latter is a bonus, not a strict requirement for the initial implementation).
\end{lstlisting}

I consider a crucial aspect of the prompt design to be the explicit definition of the class initialization and core method signatures. Without such clear specifications, a generative model is more prone to produce irrelevant outputs, as it navigates a significantly larger search space to satisfy the request. Analogous to instructing humans, constraining the output of a GenAI model necessitates clear and well-defined instructions.

Nevertheless, increasing the number of distinct concepts or complex constraints within a single prompt generally elevates the probability of model failure. Consequently, decomposing a complex request into more specific, modular prompts is often a more effective strategy for guiding GenAI. However, even this level of modularization falls short of fully addressing an LLM’s inherent struggle with the intricate logic and hierarchical complexity posed by challenges like the \texttt{AdaptiveHierarchicalTaskAssigner}. A similar situation arises with human programmers, who, depending on their experience, may only be able to tackle certain parts of the problem effectively.

Yet the question remains: how far can we constrain a prompt to fulfill a requirement without making the generated output invalid or ineffective? As Aristotle might say when discussing moral virtue in humans, when we create instructions for a GenAI model to generate code, We must ask: where lies the \textit{mesotes}---the virtuous middle ground---between being explicit and giving the model greater freedom to infer code?

\bibliography{sn-article}

\end{document}